\documentclass{article}
\usepackage{xcolor}
\usepackage{ijcai26}

\usepackage{times}
\usepackage{soul}
\usepackage{url}
\usepackage[hidelinks]{hyperref}
\usepackage[utf8]{inputenc}
\usepackage[small]{caption}
\usepackage{graphicx}
\usepackage{amsmath}
\usepackage{amsthm}
\usepackage{booktabs}
\usepackage{algorithm}
\usepackage{enumitem}
\usepackage{algorithmic}
\usepackage[switch]{lineno}
\usepackage{natbib}
\newcommand{\ignore}[1]{}

\usepackage{booktabs}
\usepackage{algorithm}
\usepackage{algorithmic}
\usepackage{newfloat}
\usepackage{listings}
\usepackage{bbding}
\usepackage{adjustbox}

\floatstyle{ruled}
\newfloat{listing}{tb}{lst}{}
\floatname{listing}{Listing}

\title{Fin-PRM: A Domain-Specialized Process Reward Model for Financial Reasoning in Large Language Models}

\author{
Jie Zhu$^{1,2}$\and
Yuanchen Zhou$^2$\and
Shuo Jiang$^{2,3}$\and
Junhui Li$^1$\thanks{Corresponding Author.}\and 
Lifan Guo$^2$\footnotemark[1]\and
\\
Feng Chen$^2$\and
Chi Zhang$^2$\\
\affiliations
$^1$School of Computer Science and Technology, Soochow University\\
$^2$Qwen DianJin Team, Alibaba Cloud Computing\\
$^3$Osaka University\\
\emails
zhujie951121@gmail.com, 
lijunhui@suda.edu.cn,  \\
\{jiangshuo.jiang, lifan.lg, betterman.chenf, edward.zhang\}@alibaba-inc.com
}

\begin{document}

\maketitle

\begin{abstract}

Process Reward Models (PRMs) supervise intermediate reasoning steps in large language models (LLMs), but existing PRMs are mainly trained on general-domain data and struggle with the structured, symbolic, and fact-sensitive nature of financial reasoning. Financial tasks require not only correct final answers but also verifiable intermediate steps grounded in domain knowledge. In this paper, we propose Fin-PRM, a domain-specialized, trajectory-aware PRM for financial reasoning that jointly models step-level correctness and trajectory-level coherence, producing binary supervision signals for both local and global reasoning quality. To support reliable supervision, we construct a high-quality financial reasoning dataset of 3K trajectories, where step- and trajectory-level labels are automatically derived from multi-source reward signals, including Monte Carlo rollouts, LLM-based evaluation, and explicit financial knowledge verification. Fin-PRM defines a unified ranking score that integrates step- and trajectory-level rewards, enabling consistent use across multiple settings. We evaluate Fin-PRM in three scenarios: (1) offline trajectory selection for supervised fine-tuning, (2) reward-guided Best-of-$N$ inference for test-time scaling, and (3) process-aware reward shaping for reinforcement learning. Experiments on financial reasoning benchmarks, including CFLUE and FinQA, show that Fin-PRM consistently outperforms general-purpose PRMs and strong baselines. \ignore{These results demonstrate the importance of domain-specialized, process-level reward modeling for expert-level financial reasoning. }Our project resources will be available at \url{https://github.com/aliyun/qwen-dianjin}.
\end{abstract}

\section{Introduction}\label{sec:introduction}
Large language models (LLMs) have demonstrated remarkable capabilities in complex reasoning tasks, motivating their increasing use in specialized domains such as finance~\citep{fingpt,dianjinr1}. However, financial reasoning tasks, including financial statement analysis, investment strategy formulation, and regulatory compliance assessment, require precision, factual correctness, and logical coherence that often exceed the capabilities of general-purpose models. In finance, even minor errors in intermediate reasoning steps can invalidate entire solutions, highlighting the need for supervision mechanisms that evaluate not just final answers but the reasoning process itself.

Process reward models (PRMs) have emerged as a promising approach for supervising intermediate reasoning. PRMs assign rewards to individual steps or full reasoning trajectories, helping models select the best responses among multiple candidates and providing scalar signals for reinforcement learning~\citep{gptstep,prmlessons,rewardingprogresss,modelsthink,Bon,reasonflux,processreinforcementrewards}. Prior work has shown PRMs to be effective in general domains, but their domain-agnostic nature limits performance in financial reasoning, where structured, symbolic, and verifiable constraints must be respected. A central challenge is the quality and interpretability of reward signals. While recent work has leveraged LLMs as automated judges~\citep{surveyllmasajudge}, such signals are often opaque and insufficiently grounded in domain knowledge. In contrast, many financial reasoning steps, such as numerical computations, rule-based deductions, and fact lookups, can be verified. To leverage this structure, we introduce knowledge-aware verification signals that aggregate reward labels in a way that is both interpretable and domain-aware.

Within this framework, we introduce Fin-PRM, a domain-specialized, trajectory-aware PRM for financial reasoning. We train Fin-PRM on a curated dataset of 3,000 step-by-step reasoning trajectories from CFLUE~\citep{cflue}, a knowledge-based Chinese financial benchmark, generated with a strong reasoning model~\citep{deepseekr1} and annotated with verifiable reward signals. Fin-PRM provides reward supervision at both step and trajectory levels, supporting three applications: offline trajectory selection for supervised fine-tuning (SFT), reward-guided Best-of-$N$ inference for test-time scaling, and process-level reward shaping for reinforcement learning.

In summary, our primary contributions include:
\begin{itemize}
\item \textbf{A High-Quality Financial Reasoning Dataset}: We construct and curate 3,000 step-by-step reasoning traces with verifiable reward annotations, providing a high-quality dataset for financial reasoning supervision.
\item \textbf{A Novel Dual-Level Training Framework}: We propose a training paradigm that integrates step-level and trajectory-level rewards, enabling PRMs to learn from multi-dimensional feedback while preserving interpretability.
\item \textbf{Comprehensive Experimental Validation}: We demonstrate the effectiveness of Fin-PRM across three scenarios, offline trajectory selection for supervised fine-tuning, reward-guided Best-of-$N$ inference at test time, and process-level reward shaping for reinforcement learning, showing significant improvements in downstream financial reasoning performance.
\end{itemize}

\section{Related Work}

\subsection{Process Reward Models} 
Process Reward Models (PRMs) have emerged as a crucial framework for providing step-level supervision and interpretable reward signals in complex reasoning tasks. State-of-the-art PRMs, exemplified by MathShepherd \citep{mathshepher}, Skywork-PRM \citep{skyworkopeno}, and Qwen2.5-Math-PRM \citep{prmlessons}, employ human-annotated supervision with synthetic reward generation to deliver evaluation capabilities across diverse reasoning domains including mathematical problem solving, scientific analysis, and programming. Recent exploratory works such as ReasonFlux-PRM \citep{reasonflux} combines both step-level and template-guided trace-level reward signals, Open-PRM \citep{openprm} leverages authoritative ORMs to reverse-engineer process-level supervision signals. In application, PRMs successfully integrated into Best-of-N sampling \citep{Bon}, offline data selection \citep{dataselect}, and online reinforcement learning for model optimization \citep{anthropic}. However, effective PRM evaluation should derive its reasoning assessment capabilities from concrete thinking trajectories rather than merely final solution correctness, and real-world vertical domain applications of PRMs impose critical requirements for deep domain knowledge mastery. Guided by these thoughts, we design a domain-socialized framework that integrates trajectory-aware evaluation with expert knowledge validation, enabling more reliable process-level assessment for finical domain. 

\subsection{Data Synthesis for Reasoning Tasks} 
High-quality data has proven fundamental to developing effective reasoning models \citep{textbooksneed}. Early approaches focused on expanded existing datasets through rule-based transformations and template-driven generation \citep{edaeasydata}. These methods improving data coverage, but lacked the sophistication required for complex reasoning task. LLMs has enabled more advanced synthesis paradigms with distillation-based approaches leveraging powerful teacher models to generate high-quality reasoning traces for training efficient student models \citep{orca}. Notable contributions include WizardMath \citep{wizardmath} synthesizes reasoning data through instruction evolution, MetaMath \citep{metamath} generates diverse problems with step-by-step solutions, and OpenThoughts \citep{openthoughts} establishes systematic methodological foundations for reasoning data synthesis through comprehensive ablation studies, providing empirical evidence for data composition principles. Considered that domain-specific applications demand specialized knowledge while maintaining reasoning quality standards, we adopt CFLUE \citep{cflue} as data source and Deepseek-R1 as teacher model to obtain reasoning trace. Recent advanced approaches employ LLM-as-a-judge for automated reward labeling, which always challenged by inherent limitations including black-box evaluation processes and insufficient reproducibility \citep{llmasajudgemtbench}. Building upon these observations, we enhance our reward annotation methodology by integrating expert-based knowledge verification mechanisms, significantly improves the observability and trustworthiness of reward signals, thereby ensuring more reliable evaluation capabilities for complex financial reasoning tasks.

\begin{figure*}[t]
    \centering 
    \includegraphics[width=0.95\textwidth]{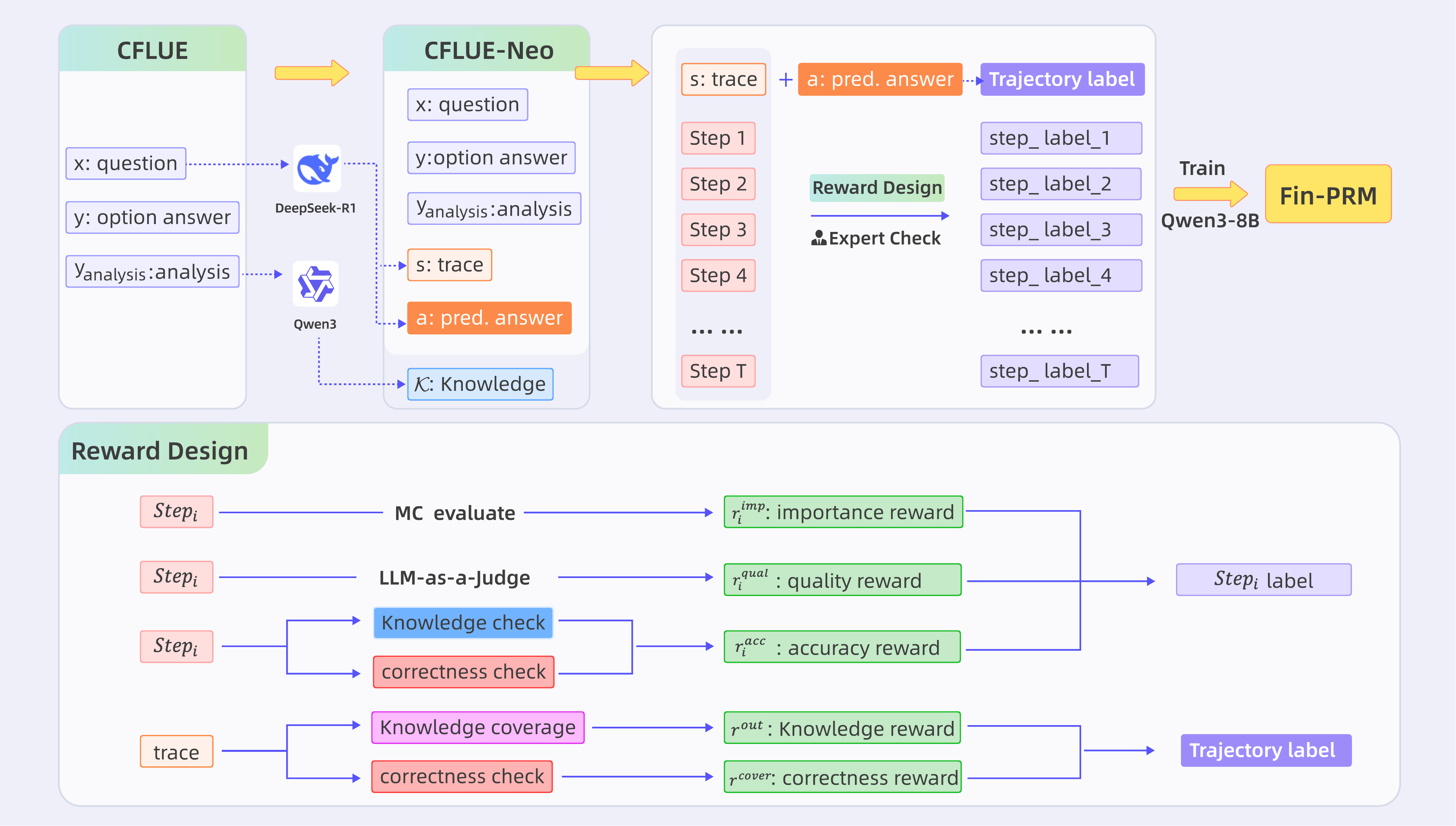}
    \caption{Illustration of the data construction process (upper) and the step-level and trajectory-level labeling procedures used for Fin-PRM training (lower).}
    \label{fig:example}
\end{figure*}

\section{Financial Reasoning Dataset Construction}\label{sec:data_construction}
The effectiveness of a process reward model (PRM) hinges on the quality of its training data~\citep{limoreasoning}. In financial reasoning, supervision must capture expert-level intermediate steps since small reasoning errors can invalidate an entire solution~\citep{gptstep}. Figure~\ref{fig:example} (upper) illustrates the Fin-PRM data construction pipeline. We next describe how we build a high-quality chain-of-thought dataset to support verifiable and interpretable reward modeling.

\subsection{Synthesizing Reasoning Trajectories}
We adopt CFLUE~\citep{cflue}, a knowledge-based Chinese financial benchmark, as our primary data source. Its Knowledge Assessment subset consists of multiple-choice questions spanning a diverse range of financial reasoning tasks and is accompanied by expert-written analyses. These analyses provide not only reliable ground-truth answers but also rich domain knowledge, making the dataset well suited for supervising and evaluating financial reasoning processes.

Following the systematic data synthesis framework of OpenThoughts~\citep{openthoughts}, we employ a strong reasoning model, Deepseek-R1~\citep{deepseekr1}, to generate structured reasoning trajectories. Given an input question $x$ from the CFLUE training split, the model produces a pair $(s, a)$, where:
\begin{itemize}[leftmargin=*]
\item $s = (s_1, s_2, \dots, s_T)$ denotes a \textit{reasoning trace}, consisting of a sequence of intermediate reasoning steps.
\item $a$ denotes the final \textit{answer} derived from the reasoning trace.
\end{itemize}
Each triplet $(x, s, a)$ forms a candidate reasoning trajectory. We first filter candidate triplets based on the length of the reasoning trace $s$ to remove overly short or degenerate trajectories. We then use Qwen3-8B to perform reasoning-on and reasoning-off evaluations for each candidate sample. A triplet is retained only if the model answers correctly with reasoning enabled but fails without reasoning. This filtering ensures that retained samples intrinsically depend on explicit reasoning rather than surface-level pattern matching.

After this preprocessing, similar to~\citet{reasonflux}, we obtain a curated dataset, CFLUE-Neo, consisting of 3,000 reasoning triplets. These triplets are then annotated with step-level and trajectory-level reward signals to train Fin-PRM.

\subsection{Financial Knowledge Extraction}
Financial reasoning is intensely knowledge-driven. To reduce reward-hacking and ensure interpretable supervision~\citep{modelsthink}, we construct an explicit financial knowledge base $\mathcal{K}$ extracted from the expert analyses provided in CFLUE.

We employ Qwen3-235b-a22b to extract key financial concepts and their definitions from expert-written explanations. For example, from an analysis discussing company valuation, the model may extract:
\begin{itemize}[leftmargin=*]
    \item \textbf{Term:} Price-to-Book Ratio
    \item \textbf{Explanation:} A financial ratio used to compare a company's current market price to its book value.
\end{itemize}
The resulting knowledge base $\mathcal{K}$ serves as an external, trusted reference for subsequent knowledge verification during reward annotation, enabling reward signals that are factually grounded rather than purely judgment-based.

Based on CFLUE, our final dataset $\mathcal{D}$ consists of tuples $(x, s, a, y, y_{\text{analysis}})$, where $y$ denotes the gold-standard answer and $y_{\text{analysis}}$ is the corresponding expert analysis. Since the teacher model may occasionally produce incorrect final answers, we treat $a$ as a silver-standard solution. In contrast, expert-provided answers, analyses, and the extracted knowledge base $\mathcal{K}$ are used as authoritative sources for verification and reward supervision.

\section{Fin-PRM: Domain-Specialized Process Reward Model}
As illustrated in the lower part of Figure~\ref{fig:example}, we define binary supervision signals at two granularities: step-level labels that capture local reasoning correctness (Section~\ref{sec:step_level}) and trajectory-level labels that reflect global reasoning validity (Section~\ref{sec:trajectory_level}). Based on these labels, we joint train Fin-PRM to predict step-wise and trajectory-wise rewards within a unified framework (Section~\ref{sec:joint_training}). 

\subsection{Step-Level Labeling}\label{sec:step_level}
Given an input question $x$ and a reasoning trajectory $s = (s_1, s_2, \dots, s_T)$, we assign a binary label to each reasoning step $s_t$, denoted as $L_{t}^{\text{step}}\in\{0, 1\}$. A positive label indicates that the step constitutes a valid and useful intermediate reasoning action, while a negative label denotes an incorrect, misleading, or unsupported step.

To determine these labels, we first compute continuous \textit{step-level reward scores} from multiple complementary perspectives, and then aggregate and binarize them to obtain final step-level labels.

\subsubsection{Step-Level Reward Scores}
To capture the multifaceted nature of high-quality financial reasoning, we decompose the step-level reward for $s_t$ into three components: an importance score $r_t^{\mathrm{imp}}$, a qualitative score $r_t^{\mathrm{qual}}$, and an accuracy score $r_t^{\mathrm{acc}}$.

\paragraph{Importance Score $r_t^{\mathrm{imp}}$.} The importance score measures the potential of a reasoning step to lead toward a correct solution. For each step $s_t$, we prompt Qwen2.5-7B-Math to generate $N$ Monte Carlo rollouts (with $N=8$ in our experiments), continuing the reasoning process from $s_t$ until a final answer is produced. The importance score is defined as the proportion of rollouts that yield a correct final answer:
\begin{equation}
\small
r_t^{\mathrm{imp}} = \frac{1}{N} \sum_{i=1}^{N} \mathbf{I}\big(\xi(\mathrm{R}{\mu,i}(s_t \mid x, s{<t}, y))\big),
\end{equation}
where $\mathrm{R}_{\mu,i}$ denotes the $i$-th rollout, $\xi(\cdot)$ is an answer-checking function, and $\mathbf{I}(\cdot)$ is the indicator function which returns 1 if the final answer of the rollout is correct and 0 otherwise. This score provides a soft estimate of whether the current step lies on a promising reasoning path.

\paragraph{Qualitative Score $r_t^{\mathrm{qual}}$.}
The qualitative score evaluates the semantic and logical quality of a reasoning step. We employ an LLM (Qwen3-235B-A22B) as an automated judge to assess each step $s_t$ with respect to semantic coherence, logical soundness, and alignment with the problem objective. The model produces a scalar score in $[0,1]$:
\begin{equation}
\small
r_t^{\mathrm{qual}} = R_{\theta}(s_t \mid x, s_{<t}, a).
\end{equation}

\paragraph{Accuracy Score $r_t^{\mathrm{acc}}$.}
The accuracy score provides supervision in verifiable ground truth and domain knowledge. It consists of two complementary components:

\textbf{Procedural Correctness ($r_t^{\mathrm{step\_corr}}$).}
This component assesses whether $s_t$ represents a logically valid and relevant step toward the gold-standard answer $y$. We prompt an LLM (Qwen3-235B-A22B) to produce a binary judgment indicating procedural correctness (1 for correct, 0 for incorrect).

\textbf{Factual Accuracy ($r_t^{\mathrm{know\_acc}}$).}
This component evaluates whether factual claims and financial concepts in $s_t$ are consistent with the extracted knowledge base $\mathcal{K}$ and expert analysis $y_{\mathrm{analysis}}$. Similarly, we prompt Qwen3-235B-A22B to produce a binary judgment indicating procedural correctness (1 for correct, 0 for incorrect). 

The final accuracy score is computed as:
\begin{equation}
\small
r_t^{\mathrm{acc}} = 0.5 \big( r_t^{\mathrm{step\_corr}}(s_t, y) + \omega_k \cdot r_t^{\mathrm{know\_acc}}(s_t, \mathcal{K}_x) \big),
\end{equation}
where $\omega_k$ controls the relative importance of factual grounding. We set $\omega_k=1.0$ in all experiments.

\subsubsection{Step-level Label Construction.}
To derive a single supervisory signal for each step, we aggregate the three reward scores using a dynamic weighting scheme based on the softmax function:
\begin{equation}
\small
r_t^{\mathrm{step}} =
\sum_{k \in {\mathrm{imp}, \mathrm{qual}, \mathrm{acc}}}
\mathrm{softmax}\big(r_t^{\mathrm{imp}}, r_t^{\mathrm{qual}}, r_t^{\mathrm{acc}}\big)_k \cdot r_t^k.
\end{equation}
This aggregation emphasizes the most informative signal among the three scores, making the supervision more robust than a fixed-weight average.

Finally, the aggregated score $r_t^{\mathrm{step}}$ is binarized using a threshold of 0.5 to obtain the step-level label:
\begin{equation}
\small
L_t^{\mathrm{step}}=\mathbf{I}\left(r_t^{\mathrm{step}} > 0.5\right).
\end{equation}

\subsection{Trajectory-Level Labeling}\label{sec:trajectory_level}
While step-level supervision encourages local correctness, a reasoning trajectory composed of individually plausible steps may still lead to an incorrect final answer. Moreover, PRMs trained solely on step-level signals are vulnerable to reward hacking, where models optimize intermediate rewards without producing valid solutions. To address these issues, we introduce trajectory-level supervision that evaluates global correctness and knowledge completeness.

\subsubsection{Trajectory-level Reward Modeling}
We define two complementary reward components for a reasoning trajectory $(s, a)$: an outcome-based correctness score $r^{\mathrm{out}}$ and a knowledge coverage score $r^{\mathrm{cover}}$.

\paragraph{Outcome Correctness Score $r^{\mathrm{out}}$.}
The outcome correctness score assesses whether the final answer $a$ is correct. Since tasks in our dataset typically require selecting a discrete option (e.g., \textit{A}, \textit{B}, \textit{ACD}), we directly compare the model's predicted answer $a$ with the gold-standard answer $y$. This yields a strict binary signal: $r^{\mathrm{out}} \in \{0, 1\}$. 

\paragraph{Knowledge Coverage Score $r^{\mathrm{cover}}$.}
Beyond correctness, high-quality financial reasoning should adequately leverage the relevant domain knowledge. The knowledge coverage score measures how thoroughly the reasoning trace and final answer utilize the required financial concepts. Formally, we define:
\begin{equation}
\small
r^{\mathrm{cover}} = \frac{|\phi_{\mathrm{ext}}(s \oplus a) \cap \mathcal{K}_x|}{|\mathcal{K}_x|},
\end{equation}
where $\mathcal{K}_x \subseteq \mathcal{K}$ denotes the subset of financial knowledge deemed relevant to the input question $x$, and $\phi_{\mathrm{ext}}(\cdot)$ extracts financial concepts mentioned in the generated reasoning and answer. The operator $\oplus$ denotes concatenation. This score encourages comprehensive reasoning grounded in the necessary domain knowledge, even when the final answer is correct.

\subsubsection{Trajectory-level Label Construction.}
We combine the two trajectory-level reward components into a single scalar score:
\begin{equation}
\small
r^{\mathrm{traj}}(s, a) = r^{\mathrm{out}}(a) + \eta \cdot r^{\mathrm{cover}}(s, a),
\end{equation}
where $\eta$ controls the relative contribution of knowledge coverage. In our experiments, we set $\eta = 1.5$, ensuring that trajectories with correct answers but insufficient knowledge grounding are penalized.

The final trajectory-level label $L^{\mathrm{traj}} \in \{0,1\}$ is obtained by thresholding $r^{\mathrm{traj}}$ at $1.25$:
\begin{equation}
\small
L^{\mathrm{traj}}=\mathbf{I}\left(r^{\mathrm{traj}} > 1.25\right).
\end{equation}

\subsection{Joint Training of Fin-PRM}\label{sec:joint_training}

Given an input question $x$, a reasoning trace $s = (s_1, \dots, s_T)$, and a final answer $a$, Fin-PRM is parameterized as a reward model $R_{\phi}$ that predicts binary reward signals at both the step and trajectory levels. In implementation, we concatenate the input question $x$, the reasoning steps $s$, and the final answer $a$ into a single sequence. A special placeholder token is inserted after each reasoning step $s_t$ and after the final answer $a$. The concatenated sequence is fed into a pretrained LLM (Qwen3-8B), and the hidden states corresponding to these placeholder tokens are extracted and passed to lightweight classification heads to produce step-level and trajectory-level predictions.

\paragraph{Step-Level Prediction.}
At the step level, Fin-PRM estimates the correctness and utility of an individual reasoning step $s_t$, conditioned on the full context:
\begin{equation}
\small
R_{\phi}^{\mathrm{step}}(s_t \mid x, s_{<t}),
\label{eq:step_reward_def}
\end{equation}
where $s_{<t}$ denotes the preceding reasoning history. This prediction reflects properties such as logical validity, numerical correctness, and relevance to the final objective.

\paragraph{Trajectory-Level Prediction.}
At the trajectory level, Fin-PRM evaluates the overall coherence, strategic soundness, and knowledge completeness of the entire reasoning process:
\begin{equation}
\small
R_{\phi}^{\mathrm{traj}}(s \mid x, a).
\label{eq:traj_reward_def}
\end{equation}
This score captures whether the reasoning trajectory follows an appropriate global strategy for solving the financial task.

\paragraph{Training Objective.}
We jointly train Fin-PRM using binary cross-entropy (BCE) loss for both step-level and trajectory-level supervision. The overall training objective is defined as:
\begin{equation}
\small
\mathcal{L}_{\mathrm{total}} = \mathcal{L}_{\mathrm{step}} + \lambda \cdot \mathcal{L}_{\mathrm{traj}},
\end{equation}
where $\lambda$ balances the contributions of step-level and trajectory-level losses.

The \textbf{step-level loss} is computed as the average BCE loss over all steps in a reasoning trace:
\begin{equation}
\small
\mathcal{L}_{\mathrm{step}} =
\frac{1}{T} \sum_{t=1}^{T}
\mathcal{L}_{\mathrm{BCE}}
\big(
R_{\phi}^{\mathrm{step}}(s_t \mid x, s_{<t}),
L_t^{\mathrm{step}}
\big),
\end{equation}
where $L_t^{\mathrm{step}} \in \{0,1\}$ denotes the ground-truth step-level label defined in Section~\ref{sec:step_level}.

The \textbf{trajectory-level loss} compares the predicted trajectory score with the corresponding trajectory-level label:
\begin{equation}
\small
\mathcal{L}_{\mathrm{traj}} = \mathcal{L}_{\mathrm{BCE}}\big(R_{\phi}^{\mathrm{traj}}(s \mid x, a), L^{\mathrm{traj}}\big),
\end{equation}
where $L_{\mathrm{traj}} \in \{0,1\}$ is constructed as described in Section~\ref{sec:trajectory_level}.

The binary cross-entropy loss is defined as:
\begin{equation}
\small
\mathcal{L}_{\mathrm{BCE}}(z, L) = -\big[L \log \sigma(z) + (1 - L)\log (1 - \sigma(z))\big],
\end{equation}
where $\sigma(\cdot)$ denotes the sigmoid function that maps model logits to probabilities.

\section{Applications of Fin-PRM}
We instantiate Fin-PRM on top of Qwen3-8B and evaluate its effectiveness across three representative application settings that reflect common usages of process reward models. In each setting, Fin-PRM is compared against relevant baselines to assess its impact on downstream financial reasoning performance.

\begin{itemize}[leftmargin=*]
\item \textbf{Offline Trajectory Selection for Supervised Fine-Tuning.} Fin-PRM is used as an offline trajectory selector to filter and curate high-quality reasoning traces, enabling more efficient and effective supervised fine-tuning (SFT).
\item \textbf{Reward-Guided Best-of-$N$ Inference.} Fin-PRM is applied at inference time to rank and select the best response from multiple candidates in a Best-of-$N$ (BoN) decoding setting.
\item \textbf{Process-Level Reward Shaping for Reinforcement Learning.} Fin-PRM serves as a dense, process-level reward function to guide policy optimization through reinforcement learning.
\end{itemize}

\ignore{To validate the effectiveness of our framework and the capability of Fin-PRM, we apply three critical use cases and compare its performance against relevant baselines.
\textbf{Supervised Fine-tuning with Data Selection:} Using Fin-PRM as an offline filter to curate a high-quality dataset for more efficient and effective SFT.
\textbf{Reward-guided Test-Time Scaling:} Employing Fin-PRM at inference time to select the best response from multiple candidates in a Best-of-N (BoN) setting.
\textbf{Online Reward Modeling:} Applying Fin-PRM as a reward function to guide the policy optimization through reinforcement learning.}

\ignore{
\begin{table*}[t]
\centering
\footnotesize
\begin{tabular}{lccccc}
\toprule
\textbf{Method} & Qwen2.5-7b-instruct & Random Selection & Math-PRM-7B & Math-PRM-72B & \textbf{Fin-PRM (Ours)} \\
\midrule
\textbf{Accuracy (\%)} & 45.3 & 43.8 & 56.5 & 57.1 & \textbf{58.2} \\
\bottomrule
\end{tabular}
\caption{Offline data selection comparison on the CFLUE benchmark. All SFT methods use 1,000 selected samples to fine-tune the Qwen2.5-7B-Instruct base model. The highest performance is in \textbf{bold}.}
\label{tab:sft_selection_horizontal}
\end{table*}
}

\begin{table}[t]
\centering
\small
\begin{tabular}{l|l}
\toprule
\textbf{Data Selection} & \textbf{Accuracy (\%)} \\
\midrule
None & 45.3\\
\midrule
Random Selection & 43.8\\
Math-PRM-7B & 56.5\\
Math-PRM-72B & 57.1\\
Fin-PRM (Ours) & \textbf{58.2}\\
\bottomrule
\end{tabular}
\caption{Accuracy of Qwen2.5-7B-Instruct on the CFLUE test set using different data selection strategies for supervised fine-tuning. Best result is shown in \textbf{bold}.}
\label{tab:sft_selection}
\end{table}

\subsection{Offline Trajectory Selection for Supervised Fine-Tuning}

PRMs can be used as offline filters to identify high-quality reasoning trajectories from large pools of synthetic data. In the financial domain, where incorrect intermediate steps can severely degrade learning, effective data selection is especially critical. In this experiment, we evaluate whether Fin-PRM can better curate supervised fine-tuning (SFT) data than general-purpose PRMs.

\paragraph{Settings.} We first use Qwen3-8B to generate multiple distinct reasoning trajectories for each question in the CFLUE training set,\footnote{To avoid coupling reward modeling with the trajectory generator, and to test generalization beyond the teacher model used for PRM training, we do not use DeepSeek-R1 at this stage.} enabling diversity in intermediate reasoning paths. Each trajectory $(x, s, a)$ is then scored by Fin-PRM using a combined step- and trajectory-level reward:
\begin{equation}
\small
\hat{R} = \frac{1}{T} \sum_{t=1}^{T} \hat{R}_{\mathrm{step}}(s_t \mid x, s_{<t}, a) + \zeta \cdot \hat{R}_{\mathrm{traj}}(s, a \mid x),
\label{eq:ranking_score}
\end{equation}
where $\hat{R}_{\mathrm{step}}$ and $\hat{R}_{\mathrm{traj}}$ are the predicted step-level and trajectory-level scores produced by Fin-PRM. The hyperparameter $\zeta$ controls the trade-off between fine-grained step correctness and overall reasoning coherence; we set $\zeta = 1.0$ in all experiments.

Based on $\hat{R}$, we select the top 1000 highest-scoring trajectories as the SFT dataset and fine-tune Qwen2.5-7B-Instruct on this curated set. We compare Fin-PRM-based selection against several baselines, including random selection and selection using Math-PRM-7B and Math-PRM-72B.

\paragraph{Results.} Table~\ref{tab:sft_selection} reports the performance on the CFLUE test set. Random data selection degrades performance from 45.3\% to 43.8\%, demonstrating the risk of training on noisy or low-quality reasoning traces. In contrast, all PRM-based selection strategies significantly outperform the base model, confirming the effectiveness of reward-guided data curation.

Among all methods, Fin-PRM achieves the best performance, reaching an accuracy of 58.2\%. This corresponds to a 12.9\% improvement over the base model and surpasses strong general-domain PRMs. These results highlight the importance of domain-specialized process reward modeling for selecting high-quality financial reasoning data.

\begin{figure}[t]
    \centering
    \includegraphics[width=1\linewidth]{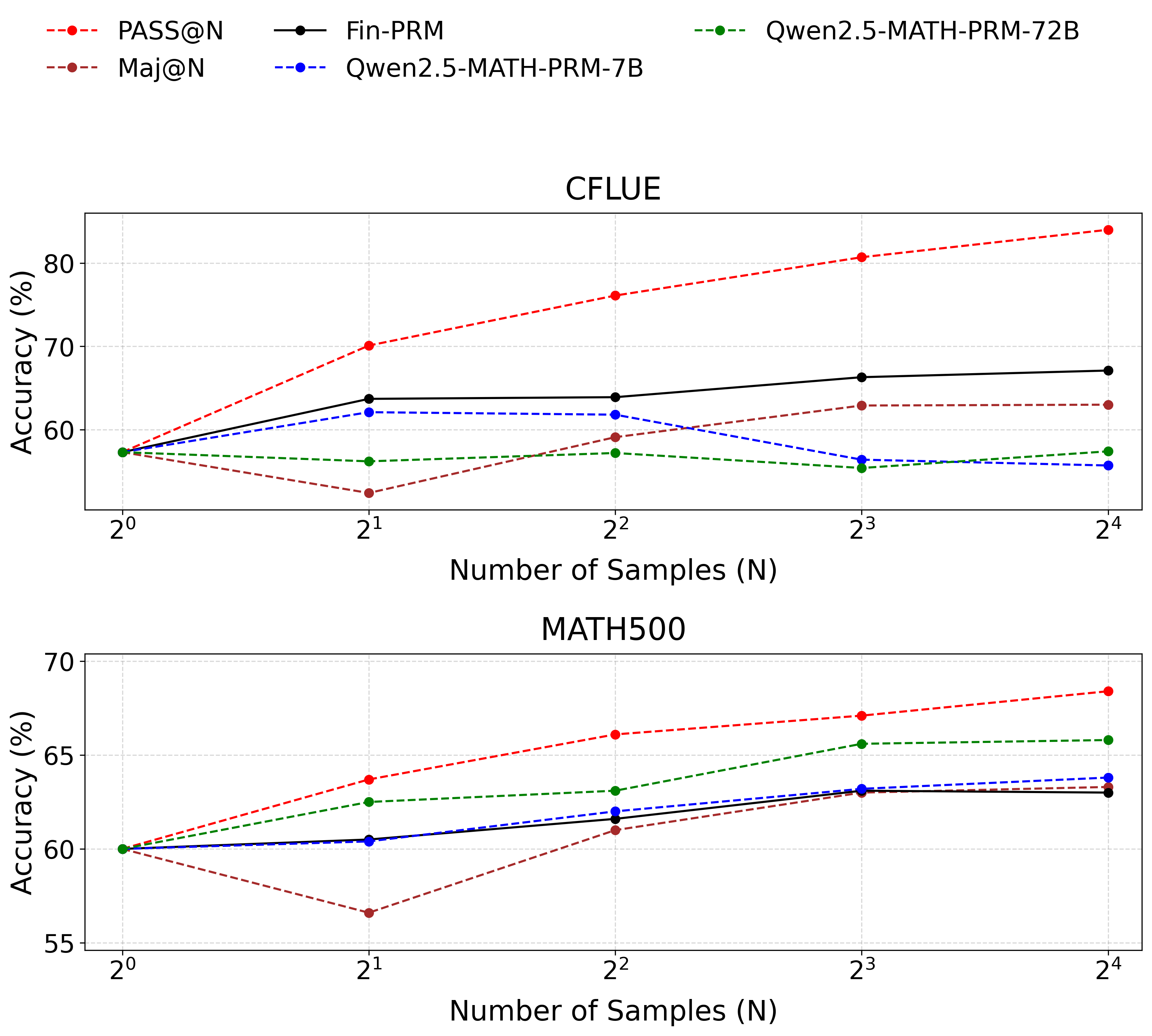}
    \caption{Best-of-$N$ selection accuracy on the CFLUE (upper) and Math500 (lower) datasets for different values of $N$.}
    \label{fig:BoN}
\end{figure}

\subsection{Reward-Guided Best-of-$N$ Inference}
A second important application of Fin-PRM is reward-guided test-time scaling via Best-of-$N$ (BoN) selection. In this setting, a policy model generates multiple candidate reasoning trajectories, and a reward model is used to select the highest-quality response. Compared with simple heuristics such as majority voting, BoN selection enables more fine-grained evaluation of reasoning quality by explicitly scoring both step-level correctness and overall trajectory quality.

\paragraph{Settings.} We evaluate BoN selection using the ranking score $\hat{R}$ defined in Eq.~\ref{eq:ranking_score}. Qwen2.5-7B-Instruct is used as the generator model to sample $N$ candidate reasoning trajectories per question. We consider $N \in \{2, 4, 8, 16\}$ and apply Fin-PRM to select the highest-scoring candidate. As baselines, we include majority voting and a strong general-domain PRMs, Qwen2.5-Math-PRM-7B and Qwen2.5-Math-PRM-72B.

We first evaluate in-domain performance on a 1,000-sample subset of the CFLUE test set, followed by an out-of-domain evaluation on the Math500 benchmark to assess generalization.

\paragraph{Results on CFLUE.} Figure~\ref{fig:BoN} upper presents the BoN results on the CFLUE dataset. Fin-PRM consistently achieves larger accuracy gains as $N$ increases, substantially outperforming majority voting across all settings. At $N=16$, Fin-PRM exceeds majority voting by more than 5.1\%, demonstrating the benefit of reward-guided selection over purely frequency-based aggregation.

In contrast, Qwen2.5-Math-PRM-7B shows diminishing returns as $N$ grows and eventually underperforms majority voting. This behavior highlights the limitations of general-domain mathematical reward models when applied to financial reasoning, and underscores the importance of domain-specific trajectory evaluation.

\ignore{
\begin{table}[t!]
\centering
\footnotesize
\begin{tabular}{lccccc}
\toprule
\textbf{Method} & \textbf{N=1} & \textbf{N=2} & \textbf{N=4} & \textbf{N=8} & \textbf{N=16} \\
\midrule
Pass@N (Oracle) & 60.0 & 63.7 & 66.1 & 67.1 & 68.4 \\
\midrule
Majority Voting & 60.0 & 56.6 & 61.0 & 63.0 & 63.3 \\
Fin-PRM (Ours) & 60.0 & 60.5 & 61.6 & 63.1 & 63.0 \\
Qwen2.5-Math-PRM-7B & 60.0 & 60.4 & 62.0 & 63.2 & 63.8 \\
Qwen2.5-Math-PRM-72B & 60.0 & \textbf{62.5} & \textbf{63.1} & \textbf{65.6} & \textbf{65.8} \\
\bottomrule
\end{tabular}
\caption{Best-of-$N$ selection accuracy on the out-of-domain Math500 benchmark. Best results are shown in \textbf{bold}.}
\label{tab:math500_bon}
\end{table}
}

\paragraph{Results on Math500.} To evaluate robustness and generalization, we conduct the same BoN experiments on the Math500 benchmark. Results are shown in the below of Figure~\ref{fig:BoN}. Fin-PRM demonstrates stable and competitive performance, outperforming majority voting for small and moderate values of $N$ (e.g., $N \leq 8$), while remaining comparable at larger $N$. Its performance closely tracks that of Qwen2.5-Math-PRM-7B at the same model scale, demonstrating that, despite being specialized for financial reasoning, Fin-PRM retains a solid generalization ability and does not overfit narrowly to the financial domain.

\subsection{Process-Level Reward Shaping for Reinforcement Learning}
Fin-PRM can also be used as a reward function to guide reinforcement learning, providing fine-grained, step-aware supervision beyond offline data selection or test-time ranking.

\paragraph{Settings.} We integrate Fin-PRM into the Group Relative Policy Optimization (GRPO) framework~\citep{grpo}, using Qwen2.5-7B-Instruct as the policy model. The evaluation is conducted on two financial reasoning benchmarks: \textit{CFLUE} and \textit{FinQA}. We compare three reward sources: 

\begin{itemize}[leftmargin=*]
\item \textbf{Outcome-only reward}: using the final-answer correctness signal $r^{\mathrm{out}}$ by comparing the predicted option answer $a$ with the gold-standard answer $y$;
\item \textbf{General-domain PRM}: Qwen2.5-Math-PRM-7B; 
\item \textbf{Fin-PRM (ours)}: the composite reward $\hat{R}$ in Eq.~\ref{eq:ranking_score} combining step-level and trajectory-level guidance.
\end{itemize}

\paragraph{Methodology.} To enable step-aware supervision, we define a composite RL reward for a reasoning trajectory $(s, a)$ as:
\begin{equation}
\small
    r_{\mathrm{rl}} = (1 - \delta) \cdot r^{\mathrm{out}} + \delta \cdot \hat{R},
\end{equation}
where $\delta$ balances the contribution of process-level reward. For a group of $N$ candidate trajectories, the advantage is computed as:
\begin{equation}
\small
    A_{\mathrm{rl}} = \frac{r_{\mathrm{rl}} - \mathrm{mean}(\{r_{\mathrm{rl}}\}_{j=1}^{N})}{\mathrm{std}(\{r_{\mathrm{rl}}\}_{j=1}^{N})}.
\end{equation}
The GRPO objective is then updated to optimize the policy with respect to $A_{\mathrm{rl}}$, including the clipping and KL-penalty terms:
\begin{equation}
\small
\label{eq:grpo_objective_refined}
\begin{array}{l}
\mathcal{J}_{\mathrm{GRPO}}(\theta) = \mathrm{E}_{x, \{s_i\} \sim \pi_{\theta_{\mathrm{old}}}} \Bigg[ \frac{1}{N} \sum_{i=1}^{N} \frac{1}{T_i} \sum_{t=1}^{T_i} \Big( \\
\quad \min \Big\{ \frac{\pi_\theta(s_{i,t} | x, s_{i,<t})}{\pi_{\theta_{\mathrm{old}}}(s_{i,t} | x, s_{i,<t})} A_{\mathrm{rl},i}, \\
\quad\quad \mathrm{clip}\big(\frac{\pi_\theta(s_{i,t} | x, s_{i,<t})}{\pi_{\theta_{\mathrm{old}}}(s_{i,t} | x, s_{i,<t})}, 1-\epsilon, 1+\epsilon\big) A_{\mathrm{rl},i} \Big\} \\
\quad - \beta_{\mathrm{KL}} \mathcal{D}_{\mathrm{KL}}(\pi_\theta \| \pi_{\mathrm{ref}}) \Big) \Bigg].
\end{array}
\end{equation}

\begin{figure}[!t]
\centering
\includegraphics[width=1\linewidth]{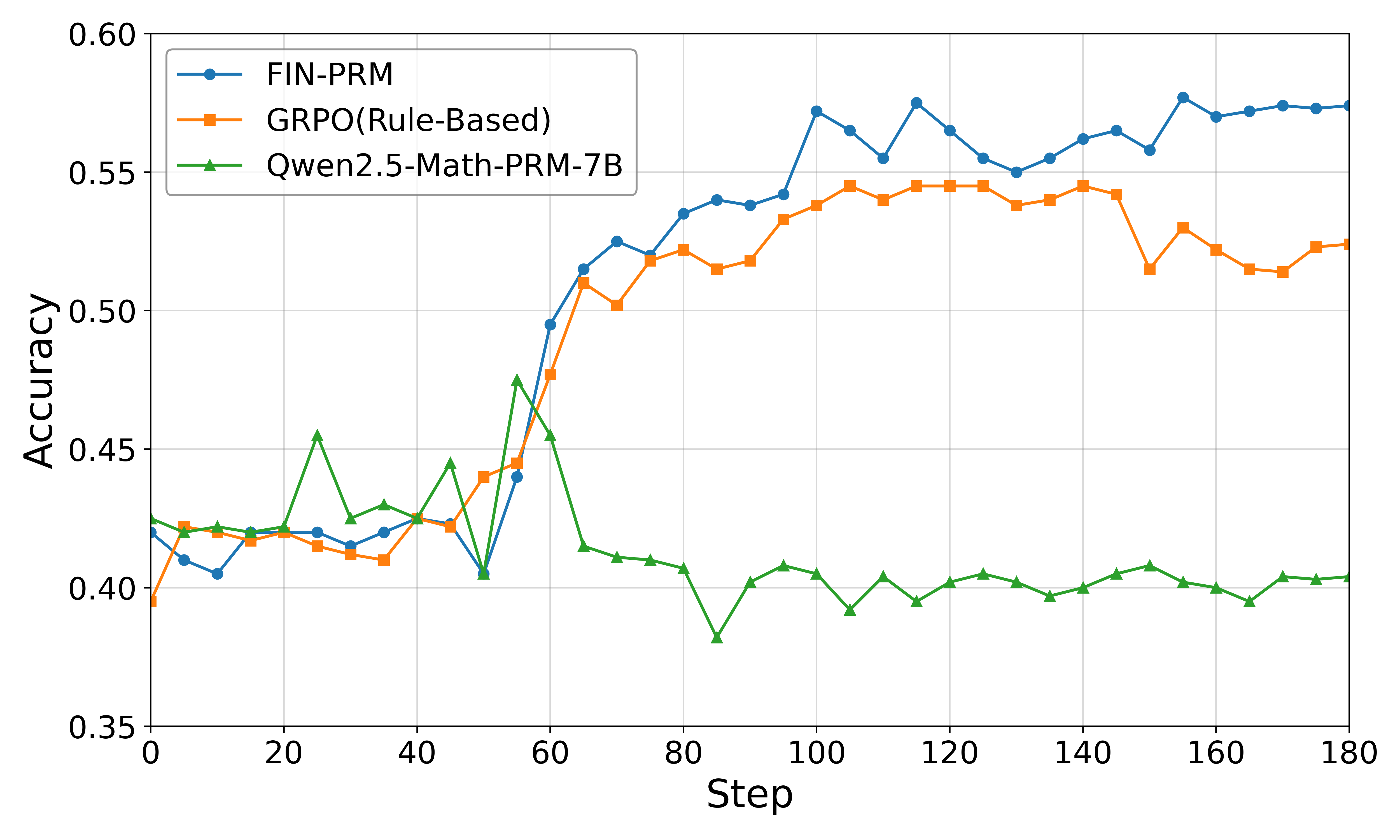}
\caption{Learning curves of GRPO-based policy optimization using different reward signals. All policies are initialized from Qwen2.5-7B-Instruct, and performance is reported as mean accuracy on the CFLUE test set over multiple runs.}
\label{fig:grpo}
\end{figure}

\paragraph{Results.} Figure~\ref{fig:grpo} shows learning curves on the CFLUE test set for policies optimized with different reward sources. In the early training stage (steps 0–60), all three strategies exhibit similar performance, suggesting comparable initial learning dynamics. In the later phase (steps 60–100), clear differences emerge: policies trained with Fin-PRM and the rule-based reward improve rapidly, while Qwen2.5-Math-PRM-7B shows shartly decrease. Throughout training, Fin-PRM consistently achieves higher accuracy and reaches the best final performance around step 100, indicating that domain-specialized, process-aware rewards provide more effective guidance than outcome-only heuristics or general-domain PRMs for financial reasoning.

Using Fin-PRM as the reward signal yields the strongest policy across evaluations, reaching an accuracy of 70.5\% on \textit{CFLUE} and 62.8\% on \textit{FinQA}. Compared to optimization with an outcome-only reward, this corresponds to an absolute improvement of 3.3\%, and it also surpasses the performance obtained with the general-domain PRM baseline. These results indicate that domain-specific, process-aware supervision provided by Fin-PRM offers more effective learning signals for reinforcement learning in financial reasoning tasks.

\ignore{
Figure \ref{fig:grpo} presents the downstream reasoning performance after using different reward signals for GRPO policy optimization. We use Qwen2.5-7B-Instruct as the policy model and compare three reward sources: a rule-based signal using only $r_{\mathrm{out}}$, the general-domain Qwen2.5-Math-PRM-7B, and our Fin-PRM.

Across all evaluations, using Fin-PRM as the reward source consistently yields the best-performing policy. Integrating Fin-PRM boosts performance on CFLUE to 70.5\% and on FinQA to 62.8\%, a significant gain of 3.3 points on both benchmarks compared to the rule-based heuristic. Crucially, Fin-PRM also outperforms the strong Qwen2.5-Math-PRM-7B baseline, highlighting that the domain-specific, factually-grounded process supervision it provides is more effective for guiding RL in a specialized domain than a general-purpose reward model. These results demonstrate that the high-quality learned reward signals from Fin-PRM substantially enhance policy optimization, leading to more capable financial reasoning models.}

\begin{figure}[t]
    \centering
    \includegraphics[width=1\linewidth]{result_Ablo.png}
    \caption{Ablation study of the ranking score weight $\zeta$. The figure reports Best-of-$N$ accuracy on the CFLUE test set as a function of $\zeta$ for different values of $N$.}
    \label{fig:Abol}
\end{figure}
\section{Ablation Study}
As defined in Eq.~\ref{eq:ranking_score}, the hyperparameter $\zeta$ controls the relative contribution of the aggregated step-level reward ($\hat{R}_{\mathrm{step}}$) and the trajectory-level reward ($\hat{R}_{\mathrm{traj}}$) when computing the final ranking score for candidate solutions. This ranking score is a core component shared across all three applications of Fin-PRM, including offline data selection, test-time Best-of-$N$ (BoN) selection, and reinforcement learning.

In this section, we use BoN selection as a representative and controlled setting to systematically study the effect of $\zeta$, as it isolates the ranking behavior of Fin-PRM without introducing additional confounding factors from model retraining.

\paragraph{Settings.} We conduct BoN selection on a 1,000-sample subset of the CFLUE test set, using the same fine-tuned Qwen2.5-7B-Instruct model as the response generator. For $N \in \{2, 4, 8, 16\}$, we vary $\zeta$ in the range $[0.0, 2.0]$ and report the resulting selection accuracy.

\paragraph{Results and Analysis.} The results are shown in Figure~\ref{fig:Abol}. Overall, performance is sensitive to the choice of $\zeta$, particularly for larger values of $N$ (i.e., $N=8$ and $N=16$). Several consistent trends emerge:

\begin{itemize}[leftmargin=*]
\item When $\zeta = 0$, the ranking relies solely on step-level rewards. While this setting yields reasonable performance, it is consistently suboptimal, indicating that local step correctness alone is insufficient for identifying the best reasoning trajectory.
\item Performance peaks around $\zeta \approx 1.0$, where step-level and trajectory-level rewards are weighted approximately equally. This suggests that effective evaluation requires a balance between fine-grained reasoning quality and global coherence.
\item As $\zeta$ increases further, accuracy gradually degrades. Overemphasizing trajectory-level scores while down-weighting step-level signals can lead to the selection of trajectories that appear globally plausible but contain critical local errors.
\end{itemize}

These results provide empirical support for our dual-granularity reward design, demonstrating that balanced integration of step-level and trajectory-level supervision is essential for effective ranking across all downstream applications of Fin-PRM. 

\section{Conclusion}
In this work, we introduce Fin-PRM, a domain-specialized, knowledge-aware process reward model that improves financial reasoning in LLMs via step- and trajectory-level supervision. Trained on 3,000 curated trajectories with verifiable reward annotations, Fin-PRM supports offline data selection, reward-guided inference, and reinforcement learning. Experiments on CFLUE and FinQA show that models trained or guided by Fin-PRM achieve significant gains, highlighting the effectiveness of domain-specific, process-aware reward modeling.

\appendix

\clearpage

\bibliographystyle{named}
\bibliography{ijcai26}

@inproceedings{cflue,
    title = "Benchmarking Large Language Models on {CFLUE} - A {C}hinese Financial Language Understanding Evaluation Dataset",
    author = "Zhu, Jie  and
      Li, Junhui  and
      Wen, Yalong  and
      Guo, Lifan",
    booktitle = "Findings of ACL",
    year = "2024",
    pages = "5673--5693"
}

@article{deepseekr1,
   title={DeepSeek-R1 incentivizes reasoning in LLMs through reinforcement learning},
   volume={645},
   number={8081},
   journal={Nature},
   author={Guo, Daya and Yang, Dejian and Zhang, Haowei and Song, Junxiao and Wang, Peiyi and others},
   year={2025},
   pages={633–638},
}

@inproceedings{mathshepher,
      title={Math-Shepherd: Verify and Reinforce LLMs Step-by-step without Human Annotations}, 
      author={Peiyi Wang and Lei Li and Zhihong Shao and R. X. Xu and Damai Dai and Yifei Li and Deli Chen and Y. Wu and Zhifang Sui},
      year={2024},
      booktitle={Proceedings of ACL},
      pages={9426-9439},
      url={https://aclanthology.org/2024.acl-long.510/}, 
}

@inproceedings{reasonflux,
      title={ReasonFlux-PRM: Trajectory-Aware PRMs for Long Chain-of-Thought Reasoning in LLMs}, 
      author={Jiaru Zou and Ling Yang and Jingwen Gu and Jiahao Qiu and Ke Shen and Jingrui He and Mengdi Wang},
      booktitle={Proceedings of NeurIPS},
      year={2025}, 
      pages={26764--26802},
      url={https://openreview.net/pdf/88d9323f85c36e29c52ce3a3cae948c2b2598eb2.pdf},
}

@inproceedings{dataselect,
      title={DoReMi: Optimizing Data Mixtures Speeds Up Language Model Pretraining}, 
      author={Sang Michael Xie and Hieu Pham and Xuanyi Dong and Nan Du and Hanxiao Liu and Yifeng Lu and Percy Liang and Quoc V. Le and Tengyu Ma and Adams Wei Yu},
      year={2023},
      pages={69798--69818},
      booktitle={Proceedings of NeurIPS},
      url={https://proceedings.neurips.cc/paper_files/paper/2023/hash/dcba6be91359358c2355cd920da3fcbd-Abstract-Conference.html}, 
}

@article{Bon,
      title={Can 1B LLM Surpass 405B LLM? Rethinking Compute-Optimal Test-Time Scaling}, 
      author={Runze Liu and Junqi Gao and Jian Zhao and Kaiyan Zhang and Xiu Li and Biqing Qi and Wanli Ouyang and Bowen Zhou},
      year={2025},
      journal={Computing Research Repository},
      url={https://arxiv.org/abs/2502.06703}, 
}

@article{skyworkopeno,
  title={Skywork-o1 Open Series},
  author={He, Jujie and Wei, Tianwen and Yan, Rui and Liu, Jiacai and Wang, Chaojie and Gan, Yimeng and Tu, Shiwen and Liu, Chris Yuhao and Zeng, Liang and Wang, Xiaokun and Wang, Boyang and Li, Yongcong and Zhang, Fuxiang and Xu, Jiacheng and An, Bo and Liu, Yang and Zhou, Yahui},
  year={2024},
  journal={Computing Research Repository},
  url={https://huggingface.co/Skywork},
}

@inproceedings{prmlessons,
  title={The Lessons of Developing Process Reward Models in Mathematical Reasoning},
  author={
    Zhenru Zhang and Chujie Zheng and Yangzhen Wu and Beichen Zhang and Runji Lin and Bowen Yu and Dayiheng Liu and Jingren Zhou and Junyang Lin
  },
  booktitle={Findings of ACL},
  year={2025},
  pages={10495-10516}
}

@inproceedings{openprm,
  title={OpenPRM: Building Open-domain Process-based Reward Models with Preference Trees},
  author={Zhang, Kaiyan and Zhang, Jiayuan and Li, Haoxin and Zhu, Xuekai and Hua, Ermo and Lv, Xingtai and Ding, Ning and Qi, Biqing and Zhou, Bowen},
  year={2025},
  booktitle={Proceedings of ICLR},
  url={https://openreview.net/pdf?id=fGIqGfmgkW},
}

@article{anthropic,
      title={Training a Helpful and Harmless Assistant with Reinforcement Learning from Human Feedback}, 
      author={Yuntao Bai and Andy Jones and Kamal Ndousse and Amanda Askell and Anna Chen and Nova DasSarma and Dawn Drain and Stanislav Fort and Deep Ganguli and Tom Henighan and Nicholas Joseph and Saurav Kadavath and Jackson Kernion and Tom Conerly and Sheer El-Showk and Nelson Elhage and Zac Hatfield-Dodds and Danny Hernandez and Tristan Hume and Scott Johnston and Shauna Kravec and Liane Lovitt and Neel Nanda and Catherine Olsson and Dario Amodei and Tom Brown and Jack Clark and Sam McCandlish and Chris Olah and Ben Mann and Jared Kaplan},
      year={2022},
      journal={Computing Research Repository},
      url={https://arxiv.org/abs/2204.05862}, 
}

@article{textbooksneed,
      title={Textbooks Are All You Need}, 
      author={Suriya Gunasekar and Yi Zhang and Jyoti Aneja and Caio César Teodoro Mendes and Allie Del Giorno and Sivakanth Gopi and Mojan Javaheripi and Piero Kauffmann and Gustavo de Rosa and Olli Saarikivi and Adil Salim and Shital Shah and Harkirat Singh Behl and Xin Wang and Sébastien Bubeck and Ronen Eldan and Adam Tauman Kalai and Yin Tat Lee and Yuanzhi Li},
      year={2023},
      journal={Computing Research Repository},
      url={https://arxiv.org/abs/2306.11644}, 
      eprint={2306.11644},
      archivePrefix={arXiv},
      primaryClass={cs.CL},
      url={https://arxiv.org/abs/2306.11644}, 
}

@inproceedings{edaeasydata,
      title={EDA: Easy Data Augmentation Techniques for Boosting Performance on Text Classification Tasks}, 
      author={Jason Wei and Kai Zou},
      year={2019},
      booktitle={Proceedings of EMNLP},
      url={https://aclanthology.org/D19-1670/}, 
      pages={6382-6388},
}

@inproceedings{wizardmath,
      title={WizardMath: Empowering Mathematical Reasoning for Large Language Models via Reinforced Evol-Instruct}, 
      author={Haipeng Luo and Qingfeng Sun and Can Xu and Pu Zhao and Jianguang Lou and Chongyang Tao and Xiubo Geng and Qingwei Lin and Shifeng Chen and Yansong Tang and Dongmei Zhang},
      year={2025},
      booktitle={Proceedings of ICLR},
      url={https://openreview.net/pdf?id=mMPMHWOdOy}, 
}

@inproceedings{metamath,
      title={MetaMath: Bootstrap Your Own Mathematical Questions for Large Language Models}, 
      author={Longhui Yu and Weisen Jiang and Han Shi and Jincheng Yu and Zhengying Liu and Yu Zhang and James T. Kwok and Zhenguo Li and Adrian Weller and Weiyang Liu},
      year={2024},
      booktitle={Proceedings of ICLR},
      url={https://openreview.net/pdf?id=N8N0hgNDRt}, 
}

@article{openthoughts,
      title={OpenThoughts: Data Recipes for Reasoning Models}, 
      author={Etash Guha and Ryan Marten and Sedrick Keh and Negin Raoof and Georgios Smyrnis and others},
      year={2025},
      journal={Computing Research Repository},
      url={https://arxiv.org/abs/2506.04178}, 
}

@article{orca,
      title={Orca: Progressive Learning from Complex Explanation Traces of GPT-4}, 
      author={Subhabrata Mukherjee and Arindam Mitra and Ganesh Jawahar and Sahaj Agarwal and Hamid Palangi and Ahmed Awadallah},
      year={2023},
      journal={Computing Research Repository},
      url={https://arxiv.org/abs/2306.02707}, 
}

@inproceedings{llmasajudgemtbench,
      title={Judging LLM-as-a-Judge with MT-Bench and Chatbot Arena}, 
      author={Lianmin Zheng and Wei-Lin Chiang and Ying Sheng and Siyuan Zhuang and Zhanghao Wu and Yonghao Zhuang and Zi Lin and Zhuohan Li and Dacheng Li and Eric P. Xing and Hao Zhang and Joseph E. Gonzalez and Ion Stoica},
      year={2023},
      pages={46595--46623},
      booktitle={Proceedings of NeurIPS},
      url={https://openreview.net/pdf?id=uccHPGDlao}, 
}

@inproceedings{rewardingprogresss,
      title={Rewarding Progress: Scaling Automated Process Verifiers for LLM Reasoning}, 
      author={Amrith Setlur and Chirag Nagpal and Adam Fisch and Xinyang Geng and Jacob Eisenstein and Rishabh Agarwal and Alekh Agarwal and Jonathan Berant and Aviral Kumar},
      year={2025},
      booktitle={Proceedings of ICLR},
      url={https://openreview.net/pdf?id=A6Y7AqlzLW},
}

@article{fingpt,
  title={FinGPT: Open-Source Financial Large Language Models},
  author={Yang, Hongyang and Liu, Xiao-Yang and Wang, Christina Dan},
  journal={FinLLM Symposium at IJCAI},
  year={2023},
}

@article{dianjinr1,
      title={DianJin-R1: Evaluating and Enhancing Financial Reasoning in Large Language Models}, 
      author={Jie Zhu and Qian Chen and Huaixia Dou and Junhui Li and Lifan Guo and Feng Chen and Chi Zhang},
      year={2025},
      journal={Computing Research Repository},
      url={https://arxiv.org/abs/2504.15716}, 
}

@inproceedings{gptstep,
      title={Let's Verify Step by Step}, 
      author={Hunter Lightman and Vineet Kosaraju and Yura Burda and Harri Edwards and Bowen Baker and Teddy Lee and Jan Leike and John Schulman and Ilya Sutskever and Karl Cobbe},
      booktitle = "Proceedings of ICLR",
      year = "2024",
      url = "https://openreview.net/pdf?id=v8L0pN6EOi",
}

@article{modelsthink,
      title={Process Reward Models That Think}, 
      author={Muhammad Khalifa and Rishabh Agarwal and Lajanugen Logeswaran and Jaekyeom Kim and Hao Peng and Moontae Lee and Honglak Lee and Lu Wang},
      year={2025},
      journal={Computing Research Repository},
      url={https://arxiv.org/abs/2504.16828}, 
}

@article{processreinforcementrewards,
      title={Process Reinforcement through Implicit Rewards}, 
      author={Ganqu Cui and Lifan Yuan and Zefan Wang and Hanbin Wang and Wendi Li and Bingxiang He and Yuchen Fan and Tianyu Yu and Qixin Xu and Weize Chen and Jiarui Yuan and Huayu Chen and Kaiyan Zhang and Xingtai Lv and Shuo Wang and Yuan Yao and Xu Han and Hao Peng and Yu Cheng and Zhiyuan Liu and Maosong Sun and Bowen Zhou and Ning Ding},
      year={2025},
      journal={Computing Research Repository},
      url={https://arxiv.org/abs/2502.01456}, 
}

@article{surveyllmasajudge,
      title={A Survey on LLM-as-a-Judge}, 
      author={Jiawei Gu and Xuhui Jiang and Zhichao Shi and Hexiang Tan and Xuehao Zhai and Chengjin Xu and Wei Li and Yinghan Shen and Shengjie Ma and Honghao Liu and Saizhuo Wang and Kun Zhang and Yuanzhuo Wang and Wen Gao and Lionel Ni and Jian Guo},
      year={2025},
      journal={Computing Research Repository},
      url={https://arxiv.org/abs/2411.15594}, 
}

@inproceedings{limoreasoning,
      title={LIMO: Less is More for Reasoning}, 
      author={Yixin Ye and Zhen Huang and Yang Xiao and Ethan Chern and Shijie Xia and Pengfei Liu},
      year={2025},
      booktitle={Proceedings of COLM},
      url={https://openreview.net/pdf?id=T2TZ0RY4Zk}, 
}

@article{grpo,
      title={DeepSeekMath: Pushing the Limits of Mathematical Reasoning in Open Language Models}, 
      author={Zhihong Shao and Peiyi Wang and Qihao Zhu and Runxin Xu and Junxiao Song and Xiao Bi and Haowei Zhang and Mingchuan Zhang and Y. K. Li and Y. Wu and Daya Guo},
      year={2024},
      journal={Computing Research Repository},
      url={https://arxiv.org/abs/2402.03300}, 
}

\clearpage

\section{Details on Models and Datasets}

\subsection{Model Implementations}

\paragraph{Generator Models.}
Throughout our experiments, we utilize several powerful large language models as generators for reasoning traces and candidate solutions:
\begin{itemize}
    \item \textbf{Deepseek-R1}: A highly capable reasoning model used in our initial data synthesis phase to generate the base reasoning trajectories from the CFLUE dataset.
    \item \textbf{Qwen3-235b-a22b}: A state-of-the-art model used for two critical auxiliary tasks: (1) extracting the structured financial knowledge base $\mathcal{K}$ from expert analyses, and (2) serving as the powerful LLM-as-a-judge to provide the qualitative score, $r_{\mathrm{qual}}$.
    \item \textbf{Qwen2.5-7B-Instruct}: A versatile and efficient model used as the base for our Fin-PRM, as well as the student model in our SFT experiments and the policy model in our GRPO and BoN experiments.
\end{itemize}

\paragraph{Reward Models (Baselines).}
To benchmark the performance of Fin-PRM, we compare it against a strong, publicly available, general-purpose PRM:
\begin{itemize}
    \item \textbf{Qwen2.5-Math-PRM-7B}: A state-of-the-art PRM specialized for the mathematics domain. It is trained on a vast corpus of math reasoning problems and serves as a powerful baseline to highlight the benefits of domain specialization. Its strong performance in a technical domain makes it a challenging benchmark for our finance-specific model.
    \item \textbf{Qwen2.5-Math-PRM-72B}: A state-of-the-art PRM specialized for the mathematics domain. The larger 72-billion parameter version, included to establish a strong upper-bound for general-purpose PRM performance on our tasks.
\end{itemize}

\subsection{Dataset Details}

\paragraph{Primary Data Source.}
Our entire framework is built upon the \textbf{CFLUE (Chinese Financial Language Understanding Evaluation)} benchmark. We selected CFLUE because it is a high-quality, knowledge-intensive dataset where questions are accompanied by detailed, human-written expert analyses. This unique feature provides the ground-truth knowledge necessary for our fact-checking reward components ($r_{\mathrm{acc}}$).

\paragraph{Synthesized Reasoning Dataset.}
We constructed a new dataset of 3,000 samples for training Fin-PRM. Each sample in our dataset, $\mathcal{D}$, is a tuple $(x, s, a, \mathcal{K}_x, y, y_{\mathrm{analysis}})$ containing:
\begin{itemize}
    \item $x$: The original question from CFLUE.
    \item $s$: The reasoning trace generated by Deepseek-R1.
    \item $a$: The final answer synthesized by Deepseek-R1.
    \item $\mathcal{K}_x$: The subset of our global knowledge base $\mathcal{K}$ relevant to question $x$.
    \item $y$: The ground-truth correct answer from CFLUE.
    \item $y_{\mathrm{analysis}}$: The ground-truth expert analysis from CFLUE, used to construct $\mathcal{K}$.
\end{itemize}
This structure enables our multi-faceted reward signal construction and provides a robust foundation for training a knowledge-aware reward model.

\subsubsection{Prompt for reasoning trace $s$ and solution $a$}
To synthesize the reasoning traces ($s$) and their corresponding solutions ($a$), we prompt the Deepseek-R1 model with a detailed set of instructions. This prompt is designed to elicit a long-form, step-by-step thought process, followed by a clean, final answer. The full prompt is shown in Listing \ref{lst:generation_prompt}. The `[Question Text]` placeholder is then replaced with the actual question from the CFLUE dataset.

\begin{listing}[h!]

\begin{lstlisting}[caption={Prompt for Generating Reasoning Traces and Solutions from Deepseek-R1}, label={lst:generation_prompt}]
Your role as an assistant involves thoroughly exploring questions through a systematic long thinking process before providing the final precise and accurate solutions. This requires engaging in a comprehensive cycle of analysis, summarizing, exploration, reassessment, reflection, backtracing, and iteration to develop a well-considered thinking process. Please structure your response into two main sections: Thought and Solution.

In the Thought section, detail your reasoning process using the specified format:
<|begin_of_thought|>
{thought with steps separated with '\n\n'}
<|end_of_thought|>
Each step should include detailed considerations such as analyzing questions, summarizing relevant findings, brainstorming new ideas, verifying the accuracy of the current steps, refining any errors, and revisiting previous steps.

In the Solution section, based on various attempts, explorations, and reflections from the Thought section, systematically present the final solution that you deem correct. The solution should maintain a logical, accurate, concise expression style and detail necessary steps needed to reach the conclusion, formatted as follows:
<|begin_of_solution|>
{final formatted, precise, and clear solution}
<|end_of_solution|>

Now, try to solve the following question through the above guidelines:
[Question Text]
\end{lstlisting}
\end{listing}
\subsubsection{Prompt for Knowledge Extraction}
To construct the ground-truth knowledge base ($\mathcal{K}_x$) for each question, we prompt the LLM judge to act as a domain expert. Its task is to read the trusted expert analysis ($y_{\mathrm{analysis}}$) provided in the dataset and extract all key financial terms along with their definitions as presented in the text. This process creates a structured, verifiable source of facts for the downstream accuracy and coverage rewards. The prompt template is detailed in Listing \ref{lst:knowledge_extraction_prompt}.

\begin{listing}[h!]
\begin{lstlisting}[caption={Template for the Knowledge Extraction Prompt}, label={lst:knowledge_extraction_prompt}]
You are a financial knowledge extraction expert. Read the following expert analysis and identify all key financial terms and concepts. For each term, provide a concise definition based on the text.

**Expert Analysis Text:**
[Expert Analysis from the Dataset]

---
**Your Task:**
Output a JSON list where each object represents a key knowledge point.

**Output Format (JSON list only):**
[
  {
    "Term": "<Identified_Term_1>",
    "Explanation": "<Definition_of_Term_1>"
  },
  {
    "Term": "<Identified_Term_2>",
    "Explanation": "<Definition_of_Term_2>"
  }
]
\end{lstlisting}
\end{listing}

\section{Details on Reward Signal Construction}
This section provides further details on the prompts used to generate the multi-faceted reward signals described in the main paper. These prompts are designed to elicit structured, machine-readable outputs from a powerful LLM judge (Qwen3-235b-a22b).

\subsection{B.1 Prompt for Qualitative Score ($r_{\mathrm{qual}}$)}
To assess the intrinsic quality of each reasoning step, we use a structured prompt that asks the LLM judge to evaluate a step ($s_t$) based on three criteria. The prompt takes the original question ($x$), the reasoning history ($s_{<t}$), and the current step ($s_t$) as input.

The LLM is instructed to provide a score from 0.0 to 1.0 for each of the following aspects, ensuring the output is a machine-parsable JSON object:
\begin{itemize}
    \item \textbf{Logical Soundness:} How coherent and logically valid is the reasoning within this specific step?
    \item \textbf{Step Correctness:} Is the information presented in the step factually or procedurally correct, independent of the overall strategy?
    \item \textbf{Target Progression:} How effectively does this step move the overall reasoning process closer to a correct final answer?
\end{itemize}
The template for this prompt is shown in Listing \ref{lst:qual_prompt}.

\begin{listing}[h!]
\begin{lstlisting}[caption={Template for the Qualitative Score Prompt}, label={lst:qual_prompt}]
You are an expert financial analyst. Given the question, the previous reasoning steps, and the current step, evaluate the quality of the **current step**.

**Question:**
[Original Question Text]

**Reasoning History:**
[Reasoning History up to step t-1]

**Current Step to Evaluate:**
[Text of step t]

---
**Your Task:**
Provide a JSON object with your evaluation based on three criteria:
1.  `logical_soundness`: How logical is the current step?
2.  `step_correctness`: Is the information in the step correct?
3.  `target_progression`: Does the step help solve the problem?

**Output Format (JSON only):**
{
  "logical_soundness": <float_from_0_to_1>,
  "step_correctness": <float_from_0_to_1>,
  "target_progression": <float_from_0_to_1>
}
\end{lstlisting}
\end{listing}
\subsection{Prompts for Verifiable and Knowledge-Based Rewards}
This subsection details the prompts used to generate rewards that are grounded in external, verifiable information, such as the ground-truth answer or our extracted knowledge base. These prompts are crucial for ensuring the factual correctness and anti-hallucination properties of Fin-PRM.

\subsubsection{Prompt for Accuracy Score ($r_{\mathrm{acc}}$)}
The accuracy score is a composite of two verifiable checks. We use distinct prompts for each to ensure a grounded, factual evaluation.

\paragraph{Procedural Correctness Prompt.} This prompt verifies if a step is a valid move towards the known correct answer ($y$), assessing its logical utility in the context of a correct solution path.

\begin{listing}[h!]
\begin{lstlisting}[caption={Template for Procedural Correctness Prompt}, label={lst:procedural_correctness_prompt}]
You are a logical verifier. Given the reasoning so far and the known correct answer, determine if the current step is a logically sound and productive move towards that answer.

**Question:**
[Original Question Text]

**Reasoning History:**
[Reasoning History up to step t-1]

**Correct Final Answer:**
[Ground Truth Answer y]

**Current Step to Evaluate:**
[Text of step t]

---
**Your Task:**
Is this step a valid, logical progression towards the correct final answer? Respond with a JSON object containing a binary value.

**Output Format (JSON only):**
{ "procedural_correctness": <1_for_yes_or_0_for_no> }
\end{lstlisting}
\end{listing}

\paragraph{Factual Accuracy Prompt.} This prompt validates the claims within a step against the extracted knowledge base ($\mathcal{K}_x$), acting as a direct anti-hallucination check.

\begin{listing}[h!]
\begin{lstlisting}[caption={Template for Factual Accuracy Prompt}, label={lst:factual_accuracy_prompt}]
You are a fact-checking agent. Verify every factual claim and financial term in the "Current Step" against the provided "Knowledge Base".

**Knowledge Base:**
- <Term_1>: <Definition_1>
- <Term_2>: <Definition_2>

**Current Step to Evaluate:**
[Text of step t]

---
**Your Task:**
Are all claims and terms in the current step supported by the knowledge base? Respond with a JSON object containing a binary value.

**Output Format (JSON only):**
{ "factual_accuracy": <1_if_all_claims_are_supported_or_0_otherwise> }
\end{lstlisting}
\end{listing}

\subsubsection{Prompt for Knowledge Coverage ($r_{\mathrm{cover}}$)}
To calculate the trajectory-level knowledge coverage score, this prompt asks the LLM to verify which of the required knowledge points ($\mathcal{K}_x$) were used in the model's full generated response ($s$ and $a$).

\begin{listing}[h!]
\begin{lstlisting}[caption={Template for the Knowledge Coverage Prompt}, label={lst:knowledge_coverage_prompt}]
You are a verification agent. Your task is to check if the required financial knowledge points were used in the provided model response.

**Required Knowledge Points:**
1. <Term_1>: <Definition_1>
2. <Term_2>: <Definition_2>
...

**Model's Reasoning Trace and Answer:**
[Model's Full Generated Response]

---
**Your Task:**
Analyze the model's response and determine how many of the required knowledge points were covered. Output a JSON object with the count and indices of the covered points.

**Output Format (JSON only):**
{
  "coverage_number": <integer>,
  "coverage_index": [<list_of_covered_indices>]
}
\end{lstlisting}
\end{listing}

\section{Additional Experimental Setups}
This section provides detailed configurations and hyperparameters for the training of Fin-PRM and its application in the three downstream tasks.

\subsection{Fin-PRM Training Details}
Fin-PRM was trained by fine-tuning the Qwen2.5-7B-Instruct model on our newly constructed financial reasoning dataset. The training objective combined step-level and trajectory-level losses, as described in Equation 16. The key hyperparameters used for training are summarized in Table \ref{tab:prm_training_hyperparams}. All training was conducted on NVIDIA A100 GPUs.

\begin{table}[h!]
\centering
\caption{Hyperparameters for Fin-PRM Training.}
\label{tab:prm_training_hyperparams}
\begin{tabular}{lc}
\toprule
\textbf{Parameter} & \textbf{Value} \\
\midrule
Base Model & Qwen2.5-7B-Instruct \\
Dataset Size & 3,000 samples \\
Learning Rate & 2e-5 \\
Batch Size & 1(per device) \\
Gradient Accumulation Steps & 2\\
Max Sequence Length & 8192 \\
Epochs & 3 \\
Optimizer & AdamW \\
LR Scheduler & Cosine with warmup \\
Warmup Steps & 50 \\
Loss Weight ($\lambda$ in Eq. 16) & 1.0 \\
\bottomrule
\end{tabular}
\end{table}

\subsection{Downstream Task Setups}
This subsection details the experimental setups for the three application scenarios: Supervised Fine-Tuning (SFT), Best-of-N (BoN) selection, and Group Relative Policy Optimization (GRPO).

\subsubsection{SFT with Data Selection}
For the offline data selection task, we first used Fin-PRM to score and select the top 1,000 reasoning traces from a larger pool of synthetic data. We then fine-tuned the Qwen2.5-7B-Instruct model on this curated subset. The SFT process used the same set of hyperparameters as the PRM training (see Table \ref{tab:prm_training_hyperparams}), ensuring a fair comparison.

\subsubsection{Best-of-N (BoN) Selection}
In the test-time scaling experiments, the policy model (Qwen2.5-7B-Instruct) generated N candidate responses for each question. Fin-PRM then scored each candidate using the composite ranking score from Equation 20. The candidate with the highest score was selected as the final answer. Based on our ablation study (Figure \ref{fig:Abol}), the hyperparameter $\zeta$ was set to 1.0 to give equal weight to step-level and trajectory-level rewards.

\subsubsection{GRPO Reinforcement Learning}
For the online policy optimization, we integrated Fin-PRM into the GRPO framework. The policy model was Qwen2.5-7B-Instruct, initialized from the same base checkpoint as other experiments. The composite reward from Equation 22 was used to calculate advantages. Key GRPO-specific hyperparameters are listed in Table \ref{tab:grpo_hyperparams}.

\begin{table}[h!]
\centering
\caption{Hyperparameters for GRPO Policy Optimization.}
\label{tab:grpo_hyperparams}
\begin{adjustbox}{width=\columnwidth}
\begin{tabular}{lc}
\toprule
\textbf{Parameter} & \textbf{Value} \\
\midrule
Policy Model & Qwen2.5-7B-Instruct \\
Samples per Group (N) & 6 \\
Reward Composition ($\delta$ in Eq. 22) & 0.5 \\
KL Penalty Coefficient ($\beta_{\mathrm{KL}}$) & 0.001 \\
PPO Clipping Parameter ($\epsilon$) & 0.2 \\
Epochs & 1 \\
Batch Size & 32 \\
Learning Rate & 1e-6 \\
\bottomrule
\end{tabular}
\end{adjustbox}
\end{table}

\section{Case Study: A Data Sample in Detail}
To better illustrate the granular nature of our reward signals, this section presents a complete data sample in a format that explicitly links each reasoning step to its corresponding reward score.

\begin{quote}
\textbf{Question:} \\
A business of a commercial bank that does not constitute on-balance-sheet assets or liabilities but generates non-interest income is called its (\hspace{1cm}).\\
\textit{Choices: A: Investment business, B: Asset business, C: Intermediary business, D: Liability business}
\end{quote}

\hrulefill 

\subsection*{Thinking Trajectories}

\paragraph{Step 1: Analyze the Problem}
The problem requires identifying a commercial bank's business type. The key criteria are: (1) It does not constitute on-balance-sheet assets or liabilities, meaning it is an off-balance-sheet business. (2) It generates non-interest income, such as fees or commissions. The options are A. Investment, B. Asset, C. Intermediary, and D. Liability business. The reasoning must be based on the classification of commercial bank operations.
\hfill \textit{[Step-level reward: 0.644]}

\paragraph{Step 2: Summarize Findings}
- \textbf{Asset business (Option B):} Involves the use of bank funds, forming on-balance-sheet assets (e.g., loans, securities investments). It primarily generates interest income and does not meet the "off-balance-sheet" condition.
- \textbf{Liability business (Option D):} Involves the bank's sources of funds, forming on-balance-sheet liabilities (e.g., deposits, borrowings). It primarily incurs interest expenses and does not fit the criteria.
- \textbf{Investment business (Option A):} Typically refers to the bank investing in financial instruments (bonds, stocks), which is a subset of asset business. It constitutes on-balance-sheet assets.
- \textbf{Intermediary business (Option C):} Refers to the bank acting as a service intermediary, not using its own funds. It is off-balance-sheet and its income comes from fees and commissions (non-interest income). This matches the description.
\hfill \textit{[Step-level reward: 0.769]}

\paragraph{Step 3: Brainstorm New Ideas}
To test the robustness of the conclusion, this step considers potential edge cases. For instance, do certain investment activities like derivatives trading count? Or do credit card fees (non-interest income) from a liability product complicate the definition? This deepens the analysis beyond a superficial match.
\hfill \textit{[Step-level reward: 0.753]}

\paragraph{Step 4: Verify Step Accuracy}
This step cross-references the definition of "intermediary business" with formal regulations, such as a country's Commercial Bank Law or accounting standards. It confirms that these formal sources align with the conclusion from Step 2, providing an external, factual grounding.
\hfill \textit{[Step-level reward: 0.686]}

\paragraph{Step 5: Refine Errors}
The model considers and clarifies potential misconceptions. For example, it distinguishes between a contingent liability (like a letter of credit, common in intermediary business) which is off-balance-sheet until triggered, and a direct on-balance-sheet liability. This refines the understanding and prevents oversimplification.
\hfill \textit{[Step-level reward: 0.839]}

\paragraph{Step 6: Revisit Previous Steps}
The final step involves a full review of the reasoning chain. It confirms that the initial analysis in Step 1 was correct, the evaluation in Step 2 was sound, and the verifications in subsequent steps all support the same conclusion, ensuring the entire trajectory is logically coherent.
\hfill \textit{[Step-level reward: 0.837]}

\hrulefill 

\subsection*{Final Response}

Based on the thought process, the business is the commercial bank's \textbf{intermediary business}.
\begin{itemize}
    \item \textbf{Reason:} Intermediary business does not constitute on-balance-sheet assets or liabilities (i.e., it is an off-balance-sheet business) but generates non-interest income (such as fees and commissions) by providing intermediary services (e.g., payment settlement, agency services, consulting).
    \item \textbf{Exclusion of other options:}
        \begin{itemize}
            \item A. Investment business: Constitutes an on-balance-sheet asset.
            \item B. Asset business: Directly forms an on-balance-sheet asset.
            \item D. Liability business: Forms an on-balance-sheet liability.
        \end{itemize}
\end{itemize}
Therefore, the correct answer is C. Intermediary Business.
\vspace{1em}
\hfill \textbf{[Trajectory-level Label: 1 (Correct)]}
\end{document}